\pdfoutput=1

\documentclass[11pt]{article}

\usepackage{acl}

\usepackage{times}
\usepackage{latexsym}

\usepackage{algorithm}
\usepackage{algpseudocode}
\usepackage{graphicx}
\usepackage{booktabs}
\usepackage{float}
\usepackage{multicol}
\usepackage{multirow}
\usepackage{amssymb}
\usepackage{xcolor}
\usepackage{pifont}
\usepackage{enumitem}
\usepackage{amsmath}

\definecolor{red}{RGB}{212,17,89}
\definecolor{green}{RGB}{91,179,24}
\definecolor{lightgreen}{RGB}{190,231,166}
\definecolor{coral}{RGB}{253,124,91}
\definecolor{navy}{RGB}{86,111,157}
\definecolor{amber}{RGB}{252,200,0}

\usepackage[T1]{fontenc}

\usepackage[utf8]{inputenc}

\usepackage{microtype}

\usepackage{inconsolata}

\title{Measuring the Inconsistency of Large Language Models\\in Preferential Ranking}
\author{Xiutian Zhao \\ University of Edinburgh \\ \texttt{x.zhao-103@sms.ed.ac.uk}
        \And  Ke Wang, Wei Peng  \\ Huawei IT Innovation and Research Center \\ \texttt{\{wangke215, peng.wei1\}@huawei.com}}

\begin{document}
\maketitle
\begin{abstract}
Despite large language models' (LLMs) recent advancements, their bias and hallucination issues persist, and their ability to offer consistent preferential rankings remains underexplored.
This study investigates the capacity of LLMs to provide consistent ordinal preferences, a crucial aspect in scenarios with dense decision space or lacking absolute answers. 
We introduce a formalization of consistency based on order theory, outlining criteria such as transitivity, asymmetry, reversibility, and independence from irrelevant alternatives. Our diagnostic experiments on selected state-of-the-art LLMs reveal their inability to meet these criteria, indicating a strong positional bias and poor transitivity, with preferences easily swayed by irrelevant alternatives. These findings highlight a significant inconsistency in LLM-generated preferential rankings, underscoring the need for further research to address these limitations.
\end{abstract}

\section{Introduction}
Expressing one's preferences in an ordinal manner is a widespread and informative practice in human reasoning and communication \cite{arrow2010handbook}. By evaluating and comparing available options, individuals can make more informed decisions and communicate their values to others more effectively. In the domain of natural language processing (NLP), human preferential feedback serves as a valuable data type for aligning language models with human inclinations \cite{schulman2017proximal, rafailov2023direct}.

\begin{figure}[t]
\centering
\includegraphics[width=0.95\columnwidth]{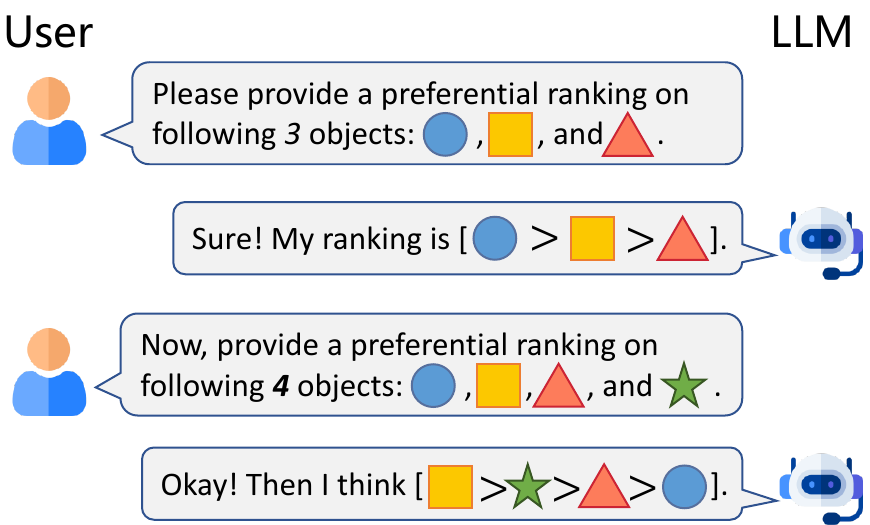}
\caption{
An example of violating \textit{Independence from Irrelevant Alternatives (IIA)} criterion. 
Initially, given 3 choices, the model preferred \textcolor{navy}{Circle} over \textcolor{amber}{Square} over \textcolor{coral}{Triangle}. However, after introducing a new choice \textcolor{green}{Star}, the \textbf{relative preferential positions} among the initial 3 choices inconsistently changed.}
\label{fig:main_art}
\end{figure}

Recent advances in large language models (LLMs) have prompted researchers to investigate the potential of LLMs in complex ranking-based tasks - such as recommendation \cite{li2023prompt, ren2024representation}, web search \cite{sun-etal-2023-chatgpt}, and text relevance comparison \cite{qin2023large} - traditionally handled by task-specific models. Moreover, given that human annotation and evaluation are resource-intensive, there is an increasing interest in augmenting or even substituting human preferential data with LLM-generated judgments to annotate, evaluate, or supplement as corpus \cite{wang2021want, zhao2022lmturk, lee2023rlaif}.

On the other hand, it is well-recognized that LLMs often exhibit severe bias and hallucination \cite{rawte2023survey, zhang2023siren}. Specifically, prior research has identified undesirable behavioral patterns in LLMs when presented with multiple options (i.e., choices). For example, in Multiple-Choice Question Answering (MCQA), a task commonly used to benchmark LLM performance \cite{hendrycks2021measuring, hendrycks2021measuringmath, robinson2023leveraging}, LLMs have shown a particular bias towards the position \cite{pezeshkpour2023large} and the labeling of choices \cite{zheng2023large}.

Unlike MCQA, which requires single-selections, preferential ranking necessitates the ordinal preferences of all options, which is invaluable in scenarios lacking a definitive answer. Despite extensive research, the current literature on LLM bias has not fully addressed their behavior in preferential ranking tasks. To address this gap, our study endeavors to investigate a critical yet under-explored question: \textit{To what extent can LLMs consistently and coherently provide ordinal `preferences'?}

This study makes an effort to measure the consistency (or more likely inconsistency) of LLMs in preferential ranking.
Firstly, by incorporating order theory \cite{gratzer2002general}, we formalize the question and define five self-evident criteria that must be satisfied to achieve `consistency' (\S~\ref{sec:definition}). Through comprehensive diagnostic experiments on various state-of-the-art LLMs, we examine their adherence to preferential ranking criteria, namely transitivity, asymmetry, reversibility, and independence from irrelevant alternatives (IIA, as exemplified in Figure~\ref{fig:main_art}). We demonstrate that even the most advanced LLMs are incapable of providing consistent or coherent preferential rankings.

Specifically, we observe that:
(1) The tested models generally \textbf{fail to meet the asymmetry condition} in preferential ranking (e.g., different answers for `compare A and B' and `compare B and A'), indicating a strong positional bias (\S~\ref{sec:asymmetry}).
(2) The preferences provided by the tested models exhibit \textbf{poor transitivity}; that is, concatenating binary preferences of choice pairs does not reliably yield an ordinal chain, and in fact, these preferences are often contested or even cyclic (\S~\ref{sec:asymmetry}).
(3) The preferences of LLMs are \textbf{significantly influenced by the addition or removal of irrelevant alternatives} (\S~\ref{sec:iia}).
(4) When requested to provide rankings in different ordinal sequences (e.g., preferential descending and ascending), LLMs \textbf{fail to produce logically equivalent outcomes} (\S~\ref{sec:reversibility}).

In summary, our contributions are threefold:
\begin{itemize}
    \item We first formalize the measurement of consistency in LLM preferential ranking through the lens of order theory.
    \item We devise specific measurement metrics that align with the defined consistency conditions. A preliminary experiment not only corroborates some shared biases with the MCQA task but also highlights the unique challenges of preferential ranking.
    \item Through comprehensive experiments on a collection of state-of-the-art (SOTA) LLMs, we uncover a severe and widespread inconsistency in LLM preferential ranking. Our findings sound a serious alarm in related research and call for immediate mitigation efforts.
\end{itemize}

\section{Experiment Setup and Preliminaries}
\subsection{Definition}\label{sec:definition}
Concretely, let $A = \{a, b,..., n\}$ be a finite set of $n$ distinct alternatives, we define a preferential ranking as a \textit{strict partial ordering} $\succ$ of $A$ \cite{gratzer2002general, rosen2007discrete}. Such ordering satisfies that, for all $a,b,c \in A$:
\begin{itemize}
    \item \textbf{Irreflexivity}: not $a > a$. 
    
    \item \textbf{Asymmetry}: if $a > b$ then not $b > a$.

    \item \textbf{Transitivity}: if $a > b$ and $b > c$ then $a > c$. 
\end{itemize}

Besides above intrinsic conditions, In multi-round preferential ranking scenarios, we also examine following criteria 

\begin{itemize}
    \item \textbf{Independent from Irrelevant Alternative (IIA)}: if $a > [...] > b$ in an ordering $\succ_{original}$, given additional alternatives ${c, d, ...}$, then $a > [...] > b$ in the new ordering $\succ_{new}$.
    
    \item \textbf{Reversibility}: if $a > [...] > b$ then $b < [...] < a$. 
    This criterion can be regarded as a full ranking generalization of binary \textit{Asymmetry}.
\end{itemize}

Note that there is also a \textit{non-strict partial ordering} variation that allows $a = b$ (i.e., preferential ties). For simplicity, all experiments are conducted under \textit{strict partial ordering} scenario.

\subsection{Datasets}
Following prior research that benchmarks the reasoning capabilities of LLMs \cite{10.1145/3526113.3545616, liu2023dynamic, zhang2023exploring, geminiteam2023gemini, jiang2023mistral}, we choose the MMLU \cite{hendrycks2021measuring} as our principal testbed. This benchmark encompasses a total of 14,079 MCQA test cases across 57 varied subject areas. Given that our study's main focus is on preference ranking rather than choice generation, the MMLU is particularly well-suited to our research interests, as the benchmark is uniformly formatted with multiple-choice options, and the options (i.e., choices) are predefined.

It should be noted that preferential ranking is a more challenging task than MCQA because it necessitates additional ordinal information. To create a balanced test set, we curate a collection by selecting the first 20 cases from each subject, resulting in a total of 1,140 cases. In line with the original MMLU framework \cite{hendrycks2021measuring}, we employ a 5-shot example prompting strategy that leverages the dataset's fixed development set.

\subsection{Evaluated Models}
To investigate the potential inconsistencies in LLM preferential ranking, we have compiled a selection of open-source models, including Llama-3-70B \cite{llama3modelcard} and Qwen-1.5-72/110B \cite{qwen1.5}. Our selection criteria prioritized models with relatively large parameters (exceeding 70B), as smaller models are generally outperformed by their larger counterparts in text-based task performance. For proprietary models, we have included gpt-3.5 \cite{brown2020language} and gpt-4o \cite{openai2023GPT4}, which are among the most widely utilized closed-source models in recent times. Specifically, we adopt the snapshot models \texttt{gpt-3.5-turbo-0125} and \texttt{gpt-4o-2024-05-13}.
Sources of tested open-source models are summarized in Table~\ref{table:models}.
Detailed specifications and sources for the selected models are provided in Appendix~\ref{appendix:reproducibility}. To ensure reproducibility, we have set the temperature for all experiments to zero (the temperature setting ranges from 0 to 2 for OpenAI models, and from 0 to 1 for others).

\subsection{Preliminary Examinations}
Prior to initiating the principal experiments, it is beneficial to ascertain whether the label tokens of choices affect LLMs's preferences and whether LLMs exhibit differential performance across single-select and preference ranking tasks.
To this end, we conduct two preliminary examinations.

\paragraph{Alternative Label Bias}
Following \cite{zheng2023large}, in comparison with the original \textit{Alphabetic} label tokens ([A, B, C, D]) of MMLU, we add \textit{Arabic}: [(1), (2), (3), (4)], and \textit{Roman}: [I, II, III, IV] token sets. The parentheses in \textit{Arabic} token to reduce ambiguity for numerical questions. Few-shot examples are modified in accordance with the altered labels.

As shown in Table~\ref{table:token}, the first-preference accuracies vary slightly for tested models. We also evaluate similarity of rankings based on minimal editing distance, and the  normalized (between 0 and 1) similarities are near 0.9, suggesting a minor influence of label tokens in preferential ranking.

\begin{table}[h]
\resizebox{\columnwidth}{!}{%
\begin{tabular}{@{}lccccc@{}}
\toprule
\multicolumn{1}{c}{\multirow{2}{*}{\textbf{\begin{tabular}[c]{@{}c@{}}Alternative\\ Labels\end{tabular}}}} &
  \textbf{Alphabetic} &
  \multicolumn{2}{c}{\textbf{Arabic}} &
  \multicolumn{2}{c}{\textbf{Roman}} \\ \cmidrule(lr){3-4} \cmidrule(lr){5-6}
\multicolumn{1}{c}{} &
  \textbf{Acc.@1} &
  \textbf{Acc.@1} &
  \textbf{Sim.} &
  \textbf{Acc.@1} &
  \textbf{Sim.} \\ \midrule
\texttt{llama-3-70b}   & 72.1 & 72.9 & 86.6 & 72.2 & 87.3 \\
\texttt{qwen-1.5-72b}  & 72.2 & 73.2 & 89.2 & 70.5 & 89.5 \\
\texttt{qwen-1.5-110b} & 71.7 & 71.9 & 90.7 & 71.1 & 89.8 \\
\texttt{gpt-3.5-turbo} & 62.1 & 62.7 & 89.1 & 61.1 & 88.0 \\
\texttt{gpt-4o}       & 83.7 & 83.1 & 92.5 & 84.6 & 92.5 \\ \bottomrule
\end{tabular}%
}
\caption{The accuracies and similarity scores of first-preferences among different label token sets. \textbf{Sim.} denotes the similarity score.
}
\label{table:token}
\end{table}

\paragraph{Question Format Sensitivity}
Given that the \textit{first-preference} in a ranking context is logically congruent with a single-select choice, we juxtapose the accuracies of MCQA across these varying question formats.

\begin{table}[h]
\centering
\resizebox{\columnwidth}{!}{%
\begin{tabular}{@{}lccccccc@{}}
\toprule
\multicolumn{1}{c}{\textbf{\begin{tabular}[c]{@{}c@{}}Question\\ Format\end{tabular}}} &
  \textbf{\begin{tabular}[c]{@{}c@{}}Single\\ Select\end{tabular}} &
  \multicolumn{3}{c}{\textbf{\begin{tabular}[c]{@{}c@{}}Ordinal\\ Ranking\end{tabular}}} &
  \multicolumn{3}{c}{\textbf{\begin{tabular}[c]{@{}c@{}}Cardinal\\ Ranking\end{tabular}}} \\ \cmidrule(lr){3-5} \cmidrule(lr){6-8}
\multicolumn{1}{c}{\textbf{HitRate@N}} &
  \textbf{-} &
  \textbf{@1} &
  \multicolumn{1}{l}{\textbf{@2}} &
  \multicolumn{1}{l}{\textbf{@3}} &
  \textbf{@1} &
  \multicolumn{1}{l}{\textbf{@2}} &
  \multicolumn{1}{l}{\textbf{@3}} \\ \midrule
\texttt{llama-3-70b}   & 77.3 & 72.1\textcolor{blue}{\small{( -5.2)}} & 85.6 & 93.1 & 73.9\textcolor{blue}{\small{( -3.4)}} & 87.4 & 94.2 \\
\texttt{qwen1.5-72b}   & 78.0 & 72.2\textcolor{blue}{\small{( -5.8)}} & 85.5 & 91.6 & 70.4\textcolor{blue}{\small{( -7.6)}} & 83.7 & 91.4 \\
\texttt{qwen1.5-110b}  & 76.5 & 71.7\textcolor{blue}{\small{( -4.8)}} & 85.9 & 93.2 & 71.8\textcolor{blue}{\small{( -4.7)}} & 86.0 & 93.2 \\
\texttt{gpt-3.5-trubo} & 67.3 & 62.1\textcolor{blue}{\small{( -5.2)}} & 81.0 & 91.6 & 61.1\textcolor{blue}{\small{( -6.2)}} & 80.7 & 91.4 \\
\texttt{gpt-4o}        & 78.1 & 83.7\textcolor{red}{\small{( +5.6)}}  & 92.2 & 96.9 & 82.7\textcolor{red}{\small{( +4.6)}}  & 92.4 & 96.8 \\ \bottomrule
\end{tabular}%
}
\caption{
Full results of Question Format Sensitivity test. \textbf{@N} denotes that the accuracies are calculated based on the first \textit{N} elements of the preferential ranking lists. 
}
\label{table:qformat}
\end{table}

\textit{Ordinal Ranking} necessitates that outputs be sequenced lists, whereas \textit{Cardinal Ranking} obliges models to assign a numerical score to each potential answer.
As evidenced in Table~\ref{table:qformat}, all models, with the exception of \texttt{gpt-4o}, demonstrate reduced accuracies relative to the single-select format. This leads us to infer that LLMs are indeed \textbf{sensitive to the format of questions}, thereby underscoring the importance of probing into LLMs' performance in preference-based ranking.

\section{Main Experiments and Key Observations}
\subsection{Irreflexivity}
We elect to forgo further scrutiny of this criterion, as a pilot test that we performed indicates that carefully crafted prompt instructions effectively preclude the recurrence of options within model-generated rankings, with infrequent transgressions observed (less than 1\% across all evaluated models).

\subsection{Asymmetry and Transitivity} \label{sec:asymmetry}
Symmetry checking in LLM reasoning is fundamentally a test for positional bias. Given that prior research has identified positional bias in single-selection tasks \cite{pezeshkpour2023large, zheng2023large}, it is imperative to ascertain whether LLMs exhibit a similar propensity to modify their preference rankings when options are sequenced differently in questions.

\begin{figure}[h]
\centering
\includegraphics[width=0.9\columnwidth]{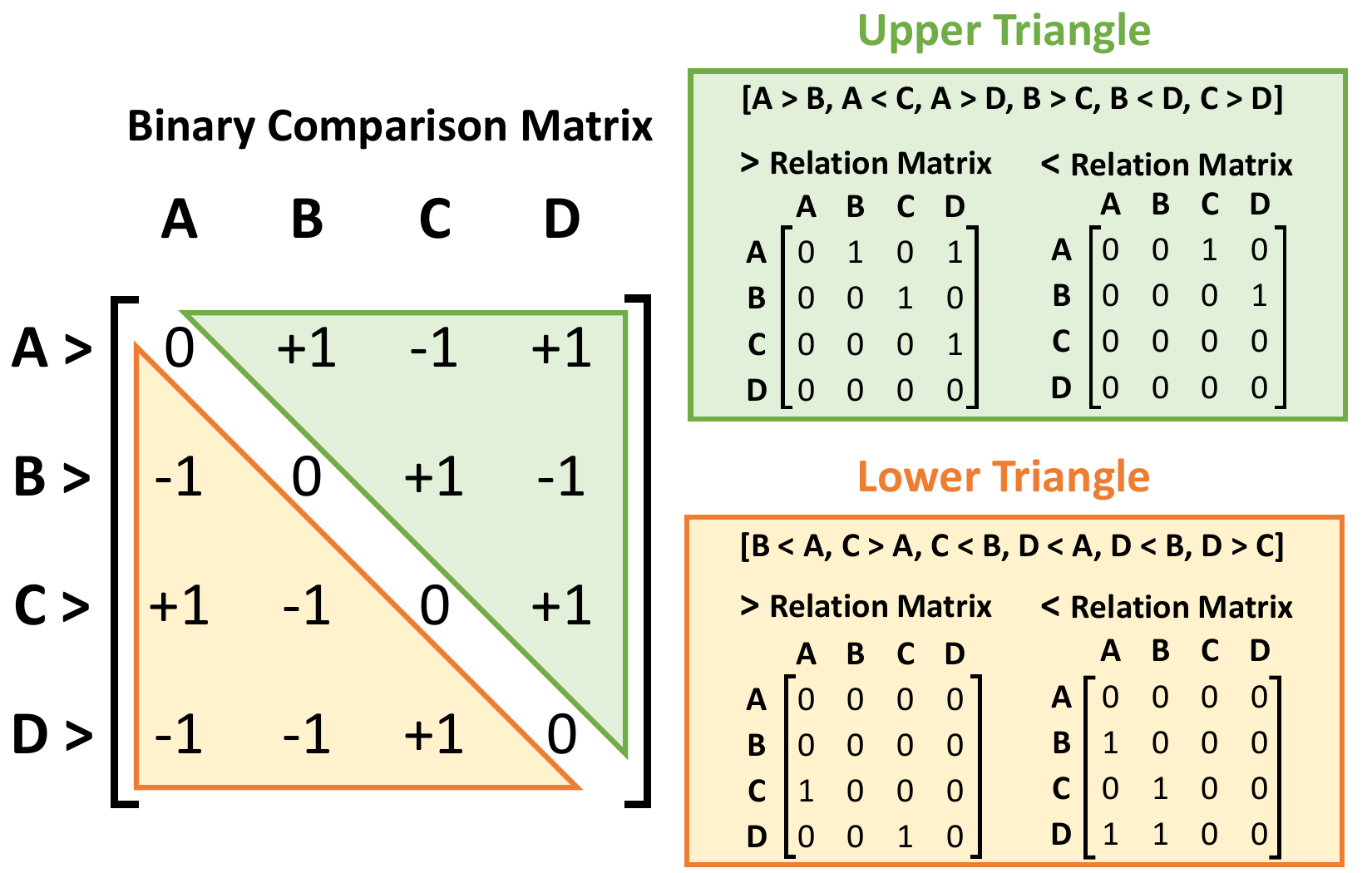}
\caption{
A 4-option binary comparison matrix (left) and a breakdown of its upper and lower triangles (right). Each triangular matrices can be transformed into a relation matrix for each relation.}
\label{fig:matrix}
\end{figure}

Consider $A$ as a list of $n$ options: $[a_1, a_2, a_3, a_4]$. By soliciting the LLM to perform binary comparisons $n \times (n-1)$ times, we can construct an $n \times n$ binary comparison matrix $M$. As depicted in Figure~\ref{fig:matrix}, we assign $m_{ij} = 1$ if the model shows a preference for $a_i$ when presented with the ordered pair $[a_i, a_j]$ (noting that $[a_j, a_i]$ constitutes a distinct ordered pair), and $m_{ij} = -1$ if the model opts for $a_j$ when faced with the same ordered pair.

Next, we calculate an asymmetry score by comparing the agreement between $m_{ij}$ and $m_{ji}$:
\begin{align}
    \frac{2 \sum_{i, j = 0, i > j}^{n} s_{ij}}{n(n-1)}, s_{ij} = \begin{cases} 0 & \text{if } m_{ij} \equiv m_{ji} \\ 1 & \text{if } m_{ij} \not\equiv m_{ji} \end{cases}
\end{align}
The average asymmetry scores, as delineated in Table~\ref{table:transitivity}, reveal a low degree of overall asymmetry among all models, indicative of significant positional biases in the preferential ranking task. Notably, \texttt{gpt-4o}, recognized as the SOTA proprietary LLM to date \cite{arenahard2024}, registers the lowest asymmetry score. Fundamentally, the positions of options in binary comparisons markedly affects the preferences of LLMs, culminating in a decrease in duly asymmetry. This finding also concurs prior observations that LLMs show a position bias in MCQA task \cite{robinson2023leveraging}.

\begin{table}[h]
\centering
\resizebox{0.9\columnwidth}{!}{%
\begin{tabular}{@{}lcccc@{}}
\toprule
 & \multirow{2}{*}{\textbf{\begin{tabular}[c]{@{}c@{}}Asym\\ -metry\end{tabular}}} & \multicolumn{3}{c}{\textbf{Transitivity}} \\ \cmidrule(l){3-5} 
\textbf{Model}         &                           & \textbf{Upper Tri.} & \textbf{Lower Tri.} & \textbf{Avg.} \\ \midrule
random                 & \multicolumn{1}{c|}{49.9} & 59.4                & 59.4                & 59.4          \\
\texttt{llama-3-70b}   & \multicolumn{1}{c|}{76.6} & 94.5                & 94.7                & 94.6          \\
\texttt{qwen-1.5-72b}  & \multicolumn{1}{c|}{73.4} & 96.5                & 96.1                & 96.3          \\
\texttt{qwen-1.5-110b} & \multicolumn{1}{c|}{82.8} & 97.3                & 96.4                & 96.9          \\
\texttt{gpt-3.5-turbo} & \multicolumn{1}{c|}{73.0} & 94.1                & 94.6                & 94.4          \\
\texttt{gpt-4o}        & \multicolumn{1}{c|}{67.1} & 89.2                & 88.9                & 89.1          \\ \bottomrule
\end{tabular}%
}
\caption{Asymmetry and transitivity scores comparisons.
\textbf{Upper Tri.} and \textbf{Lower Tri.} denotes the upper triangle and lower triangle results, respectively.
}
\label{table:transitivity}
\end{table}

Upon identifying inconsistencies in asymmetry, we recognize that the upper and lower triangles of a binary comparison matrix do not perfectly correspond. Consequently, it is imperative to calculate transitivity separately for each triangular matrix.

Considering options as nodes and relations (`<' and `>') as directions within a graph, we reconceptualize the problem as one of directed reachability. For each relation, a relation matrix $R$ can be derived from the triangular matrices, as depicted in Figure~\ref{fig:matrix}. Subsequently, we can compute the \textit{transitive closure} matrix \cite{purdom1970transitive, karp1990transitive}:
\begin{align}
    M_t = [r_{ij}]_{n \times n} = R^0 \vee R^1 \vee ... \vee R^{n-1}
\end{align}
where $\vee$ is \textit{logical and} operation. 
If $r_{ij} = 1$ in $M_t$, then the relation has successfully transitioned; otherwise, it is deemed non-transitive.

As evidenced in Table~\ref{table:transitivity}, all models exhibit moderate transitivity, with the random baseline established at 59.4. 
Furthermore, there are subtle differences between the upper and lower triangles in all models. This observation is consistent with the positional biases identified in the asymmetry experiment. In contrast, the impact of relation symbols on transitivity is considerably less pronounced.

\subsection{Independent from Irrelevant Alternative} \label{sec:iia}
IIA criterion assesses whether the introduction of an additional option affects the relative order of the original preference ranking.
This condition is tested by calculating a normalized similarity score, $Sim =  1- {MED}/{2n}$, where MED represents the minimum edit distance between (1) preference rankings with three options and (2) preference rankings with four options, excluding the omitted choice from the three option rankings.

\begin{table}[t]
\centering
\resizebox{0.9\columnwidth}{!}{%
\begin{tabular}{@{}lccccc@{}}
\toprule
\multicolumn{1}{c}{\textbf{\begin{tabular}[c]{@{}c@{}}Removed\\ Choice Index\end{tabular}}} &
  \textbf{Gold} &
  \textbf{\begin{tabular}[c]{@{}c@{}}Gold\\ + 1\end{tabular}} &
  \textbf{\begin{tabular}[c]{@{}c@{}}Gold\\ + 2\end{tabular}} &
  \textbf{\begin{tabular}[c]{@{}c@{}}Gold\\ + 3\end{tabular}} &
  \textbf{\begin{tabular}[c]{@{}c@{}}Random\\ Non-Gold\end{tabular}} \\ \midrule
\texttt{llama-3-70b}   & 49.7 & 65.3 & 64.9 & 67.1 & 66.3 \\
\texttt{qwen-1.5-72b}  & 55.5 & 75.3 & 75.6 & 74.1 & 75.2 \\
\texttt{qwen-1.5-110b} & 57.9 & 76.7 & 76.3 & 76.3 & 74.6 \\
\texttt{gpt-3.5-turbo} & 62.5 & 71.5 & 71.4 & 70.6 & 69.7 \\
\texttt{gpt-4o}       & 65.9 & 80.4 & 79.7 & 80.5 & 81.3 \\ \bottomrule
\end{tabular}%
}
\caption{Similarity scores are calculated comparing with full-option rankings.
\textbf{+N} denotes the N-th option after the indices of gold answers.
}
\label{table:iia}
\end{table}

As suggested by Table~\ref{table:iia}, the removal of gold answers significantly alters the LLMs' preferences for the remaining options. Conversely, the elimination of non-gold options results in less pronounced, yet still noticeable, impacts on the preference rankings.

\subsection{Reversibility} \label{sec:reversibility}
In all preceding experiments, the models were instructed to provide preferences in a descending order, placing the most favored option first. Maintaining all other conditions constant, we now instruct the models to rank in an ascending order, positioning the least favored option at the forefront.

Table~\ref{table:reverse} encapsulates the \textit{first-N option match rates} and the overall ranking similarities between the original rankings and the reversed sequences under the alternative output order. All models exhibit suboptimal performance on full match length (with <45\% match rate), while \texttt{gpt-4o} outperforms other models by significant margins.

\begin{table}[h]
\centering
\resizebox{0.8\columnwidth}{!}{%
\begin{tabular}{@{}lcrc|c@{}}
\toprule
\multicolumn{1}{c}{\textbf{Match Length}} & \textbf{1} & \multicolumn{1}{c}{\textbf{2}} & \textbf{3 (also 4)} & \textbf{Sim.} \\ \midrule
\texttt{llama-3-70b}   & 73.4 & 47.8 & 34.1 & 80.9 \\
\texttt{qwen-1.5-72b}  & 70.1 & 42.0 & 30.3 & 79.9 \\
\texttt{qwen-1.5-110b} & 70.7 & 45.7 & 31.5 & 80.3 \\
\texttt{gpt-3.5-turbo} & 61.4 & 37.4 & 28.3 & 78.4 \\
\texttt{gpt-4o}       & 85.4 & 61.3 & 44.6 & 84.8 \\ \bottomrule
\end{tabular}%
}
\caption{Since repetitive entries are forbidden (see \S~\ref{sec:definition}), results for match length of 3 and 4 are the same for 4-option sequences. \textit{Sim.} denotes similarity scores.
}
\label{table:reverse}
\end{table}

\section{Conclusion and Future Work}\label{sec:conclusion}
To conclude, we have formalized the consistency measurements in preferential ranking tasks by designing corresponding criteria and metrics. Through diagnostic experiments, we have evaluated some of the most advanced LLMs, uncovering severe inconsistencies and positional biases that are prevalent across all models, among other observations.
Our study raises general awareness of discrepancies in LLMs and signals a call for future research efforts. Specifically, we highlight two areas of interest: the development of a non-MCQA benchmark for consistency measurement and the creation of mitigation methods to enhance the consistency of LLMs in ranking-based tasks.

\section*{Limitations}\label{sec:limitations}
While the experiments on MMLU yield notable and insightful observations, we acknowledge that MCQA is not fully aligned with preferential ranking. Most QA benchmarks have predetermined `correct' answers; however, preferential ranking can also be relevant in scenarios where there is no absolute right or wrong. Therefore, an additional avenue for future work could involve constructing a benchmark that measures preference representativeness rather than one based on true-or-false judgments.

\section*{Acknowledgements}
Our colleague Jack provided the deployment and inference services of the tested open-source models, thanks to him.

\bibliography{acl_latex}

\begin{thebibliography}{31}
\expandafter\ifx\csname natexlab\endcsname\relax\def\natexlab#1{#1}\fi

\bibitem[{AI@Meta(2024)}]{llama3modelcard}
AI@Meta. 2024.
\newblock \href {https://github.com/meta-llama/llama3/blob/main/MODEL_CARD.md} {Llama 3 model card}.

\bibitem[{Arrow et~al.(2010)Arrow, Sen, and Suzumura}]{arrow2010handbook}
Kenneth~J Arrow, Amartya Sen, and Kotaro Suzumura. 2010.
\newblock \emph{Handbook of social choice and welfare}, volume~2.
\newblock Elsevier.

\bibitem[{Brown et~al.(2020)Brown, Mann, Ryder, Subbiah, Kaplan, Dhariwal, Neelakantan, Shyam, Sastry, Askell, Agarwal, Herbert-Voss, Krueger, Henighan, Child, Ramesh, Ziegler, Wu, Winter, Hesse, Chen, Sigler, Litwin, Gray, Chess, Clark, Berner, McCandlish, Radford, Sutskever, and Amodei}]{brown2020language}
Tom~B. Brown, Benjamin Mann, Nick Ryder, Melanie Subbiah, Jared Kaplan, Prafulla Dhariwal, Arvind Neelakantan, Pranav Shyam, Girish Sastry, Amanda Askell, Sandhini Agarwal, Ariel Herbert-Voss, Gretchen Krueger, Tom Henighan, Rewon Child, Aditya Ramesh, Daniel~M. Ziegler, Jeffrey Wu, Clemens Winter, Christopher Hesse, Mark Chen, Eric Sigler, Mateusz Litwin, Scott Gray, Benjamin Chess, Jack Clark, Christopher Berner, Sam McCandlish, Alec Radford, Ilya Sutskever, and Dario Amodei. 2020.
\newblock \href {http://arxiv.org/abs/2005.14165} {Language models are few-shot learners}.

\bibitem[{Google(2023)}]{geminiteam2023gemini}
Google. 2023.
\newblock \href {http://arxiv.org/abs/2312.11805} {Gemini: A family of highly capable multimodal models}.

\bibitem[{Gr{\"a}tzer(2002)}]{gratzer2002general}
George Gr{\"a}tzer. 2002.
\newblock \emph{General lattice theory}.
\newblock Springer Science \& Business Media.

\bibitem[{Hendrycks et~al.(2021{\natexlab{a}})Hendrycks, Burns, Basart, Zou, Mazeika, Song, and Steinhardt}]{hendrycks2021measuring}
Dan Hendrycks, Collin Burns, Steven Basart, Andy Zou, Mantas Mazeika, Dawn Song, and Jacob Steinhardt. 2021{\natexlab{a}}.
\newblock \href {http://arxiv.org/abs/2009.03300} {Measuring massive multitask language understanding}.

\bibitem[{Hendrycks et~al.(2021{\natexlab{b}})Hendrycks, Burns, Kadavath, Arora, Basart, Tang, Song, and Steinhardt}]{hendrycks2021measuringmath}
Dan Hendrycks, Collin Burns, Saurav Kadavath, Akul Arora, Steven Basart, Eric Tang, Dawn Song, and Jacob Steinhardt. 2021{\natexlab{b}}.
\newblock \href {http://arxiv.org/abs/2103.03874} {Measuring mathematical problem solving with the math dataset}.

\bibitem[{Jiang et~al.(2023)Jiang, Sablayrolles, Mensch, Bamford, Chaplot, de~las Casas, Bressand, Lengyel, Lample, Saulnier, Lavaud, Lachaux, Stock, Scao, Lavril, Wang, Lacroix, and Sayed}]{jiang2023mistral}
Albert~Q. Jiang, Alexandre Sablayrolles, Arthur Mensch, Chris Bamford, Devendra~Singh Chaplot, Diego de~las Casas, Florian Bressand, Gianna Lengyel, Guillaume Lample, Lucile Saulnier, Lélio~Renard Lavaud, Marie-Anne Lachaux, Pierre Stock, Teven~Le Scao, Thibaut Lavril, Thomas Wang, Timothée Lacroix, and William~El Sayed. 2023.
\newblock \href {http://arxiv.org/abs/2310.06825} {Mistral 7b}.

\bibitem[{Karp(1990)}]{karp1990transitive}
Richard~M Karp. 1990.
\newblock The transitive closure of a random digraph.
\newblock \emph{Random Structures \& Algorithms}, 1(1):73--93.

\bibitem[{Lee et~al.(2023)Lee, Phatale, Mansoor, Mesnard, Ferret, Lu, Bishop, Hall, Carbune, Rastogi, and Prakash}]{lee2023rlaif}
Harrison Lee, Samrat Phatale, Hassan Mansoor, Thomas Mesnard, Johan Ferret, Kellie Lu, Colton Bishop, Ethan Hall, Victor Carbune, Abhinav Rastogi, and Sushant Prakash. 2023.
\newblock \href {http://arxiv.org/abs/2309.00267} {Rlaif: Scaling reinforcement learning from human feedback with ai feedback}.

\bibitem[{Li et~al.(2023)Li, Zhang, and Chen}]{li2023prompt}
Lei Li, Yongfeng Zhang, and Li~Chen. 2023.
\newblock Prompt distillation for efficient llm-based recommendation.
\newblock In \emph{Proceedings of the 32nd ACM International Conference on Information and Knowledge Management}, pages 1348--1357.

\bibitem[{Li et~al.(2024)Li, Chiang, Frick, Dunlap, Zhu, Gonzalez, and Stoica}]{arenahard2024}
Tianle Li, Wei-Lin Chiang, Evan Frick, Lisa Dunlap, Banghua Zhu, Joseph~E. Gonzalez, and Ion Stoica. 2024.
\newblock \href {https://lmsys.org/blog/2024-04-19-arena-hard/} {From live data to high-quality benchmarks: The arena-hard pipeline}.

\bibitem[{Liu et~al.(2023)Liu, Zhang, Li, Liu, and Yang}]{liu2023dynamic}
Zijun Liu, Yanzhe Zhang, Peng Li, Yang Liu, and Diyi Yang. 2023.
\newblock \href {http://arxiv.org/abs/2310.02170} {Dynamic llm-agent network: An llm-agent collaboration framework with agent team optimization}.

\bibitem[{OpenAI(2023)}]{openai2023GPT4}
OpenAI. 2023.
\newblock \href {http://arxiv.org/abs/2303.08774} {Gpt-4 technical report}.

\bibitem[{Park et~al.(2022)Park, Popowski, Cai, Morris, Liang, and Bernstein}]{10.1145/3526113.3545616}
Joon~Sung Park, Lindsay Popowski, Carrie Cai, Meredith~Ringel Morris, Percy Liang, and Michael~S. Bernstein. 2022.
\newblock \href {https://doi.org/10.1145/3526113.3545616} {Social simulacra: Creating populated prototypes for social computing systems}.
\newblock In \emph{Proceedings of the 35th Annual ACM Symposium on User Interface Software and Technology}, UIST '22, New York, NY, USA. Association for Computing Machinery.

\bibitem[{Pezeshkpour and Hruschka(2023)}]{pezeshkpour2023large}
Pouya Pezeshkpour and Estevam Hruschka. 2023.
\newblock Large language models sensitivity to the order of options in multiple-choice questions.
\newblock \emph{arXiv preprint arXiv:2308.11483}.

\bibitem[{Purdom~Jr(1970)}]{purdom1970transitive}
Paul Purdom~Jr. 1970.
\newblock A transitive closure algorithm.
\newblock \emph{BIT Numerical Mathematics}, 10(1):76--94.

\bibitem[{Qin et~al.(2023)Qin, Jagerman, Hui, Zhuang, Wu, Shen, Liu, Liu, Metzler, Wang et~al.}]{qin2023large}
Zhen Qin, Rolf Jagerman, Kai Hui, Honglei Zhuang, Junru Wu, Jiaming Shen, Tianqi Liu, Jialu Liu, Donald Metzler, Xuanhui Wang, et~al. 2023.
\newblock Large language models are effective text rankers with pairwise ranking prompting.
\newblock \emph{arXiv preprint arXiv:2306.17563}.

\bibitem[{Qwen(2024)}]{qwen1.5}
Qwen. 2024.
\newblock \href {https://qwenlm.github.io/blog/qwen1.5/} {Introducing qwen1.5}.

\bibitem[{Rafailov et~al.(2023)Rafailov, Sharma, Mitchell, Ermon, Manning, and Finn}]{rafailov2023direct}
Rafael Rafailov, Archit Sharma, Eric Mitchell, Stefano Ermon, Christopher~D. Manning, and Chelsea Finn. 2023.
\newblock \href {http://arxiv.org/abs/2305.18290} {Direct preference optimization: Your language model is secretly a reward model}.

\bibitem[{Rawte et~al.(2023)Rawte, Sheth, and Das}]{rawte2023survey}
Vipula Rawte, Amit Sheth, and Amitava Das. 2023.
\newblock A survey of hallucination in large foundation models.
\newblock \emph{arXiv preprint arXiv:2309.05922}.

\bibitem[{Ren et~al.(2024)Ren, Wei, Xia, Su, Cheng, Wang, Yin, and Huang}]{ren2024representation}
Xubin Ren, Wei Wei, Lianghao Xia, Lixin Su, Suqi Cheng, Junfeng Wang, Dawei Yin, and Chao Huang. 2024.
\newblock Representation learning with large language models for recommendation.
\newblock In \emph{Proceedings of the ACM on Web Conference 2024}, pages 3464--3475.

\bibitem[{Robinson et~al.(2023)Robinson, Rytting, and Wingate}]{robinson2023leveraging}
Joshua Robinson, Christopher~Michael Rytting, and David Wingate. 2023.
\newblock \href {http://arxiv.org/abs/2210.12353} {Leveraging large language models for multiple choice question answering}.

\bibitem[{Rosen(2007)}]{rosen2007discrete}
Kenneth~H Rosen. 2007.
\newblock \emph{Discrete mathematics and its applications}.
\newblock The McGraw Hill Companies,.

\bibitem[{Schulman et~al.(2017)Schulman, Wolski, Dhariwal, Radford, and Klimov}]{schulman2017proximal}
John Schulman, Filip Wolski, Prafulla Dhariwal, Alec Radford, and Oleg Klimov. 2017.
\newblock \href {http://arxiv.org/abs/1707.06347} {Proximal policy optimization algorithms}.

\bibitem[{Sun et~al.(2023)Sun, Yan, Ma, Wang, Ren, Chen, Yin, and Ren}]{sun-etal-2023-chatgpt}
Weiwei Sun, Lingyong Yan, Xinyu Ma, Shuaiqiang Wang, Pengjie Ren, Zhumin Chen, Dawei Yin, and Zhaochun Ren. 2023.
\newblock \href {https://doi.org/10.18653/v1/2023.emnlp-main.923} {Is {C}hat{GPT} good at search? investigating large language models as re-ranking agents}.
\newblock In \emph{Proceedings of the 2023 Conference on Empirical Methods in Natural Language Processing}, pages 14918--14937, Singapore. Association for Computational Linguistics.

\bibitem[{Wang et~al.(2021)Wang, Liu, Xu, Zhu, and Zeng}]{wang2021want}
Shuohang Wang, Yang Liu, Yichong Xu, Chenguang Zhu, and Michael Zeng. 2021.
\newblock Want to reduce labeling cost? gpt-3 can help.
\newblock In \emph{Findings of the Association for Computational Linguistics: EMNLP 2021}, pages 4195--4205.

\bibitem[{Zhang et~al.(2023{\natexlab{a}})Zhang, Xu, and Deng}]{zhang2023exploring}
Jintian Zhang, Xin Xu, and Shumin Deng. 2023{\natexlab{a}}.
\newblock Exploring collaboration mechanisms for llm agents: A social psychology view.
\newblock \emph{arXiv preprint arXiv:2310.02124}.

\bibitem[{Zhang et~al.(2023{\natexlab{b}})Zhang, Li, Cui, Cai, Liu, Fu, Huang, Zhao, Zhang, Chen et~al.}]{zhang2023siren}
Yue Zhang, Yafu Li, Leyang Cui, Deng Cai, Lemao Liu, Tingchen Fu, Xinting Huang, Enbo Zhao, Yu~Zhang, Yulong Chen, et~al. 2023{\natexlab{b}}.
\newblock Siren's song in the ai ocean: a survey on hallucination in large language models.
\newblock \emph{arXiv preprint arXiv:2309.01219}.

\bibitem[{Zhao et~al.(2022)Zhao, Mi, Wang, Li, Jiang, Liu, and Sch{\"u}tze}]{zhao2022lmturk}
Mengjie Zhao, Fei Mi, Yasheng Wang, Minglei Li, Xin Jiang, Qun Liu, and Hinrich Sch{\"u}tze. 2022.
\newblock Lmturk: Few-shot learners as crowdsourcing workers in a language-model-as-a-service framework.
\newblock In \emph{Findings of the Association for Computational Linguistics: NAACL 2022}, pages 675--692.

\bibitem[{Zheng et~al.(2023)Zheng, Zhou, Meng, Zhou, and Huang}]{zheng2023large}
Chujie Zheng, Hao Zhou, Fandong Meng, Jie Zhou, and Minlie Huang. 2023.
\newblock Large language models are not robust multiple choice selectors.
\newblock In \emph{The Twelfth International Conference on Learning Representations}.

\end{thebibliography}

\appendix
\addcontentsline{toc}{section}{Appendices}
\renewcommand{\thesubsection}{\Alph{subsection}}

\section{Reproducibility}\label{appendix:reproducibility}
The source for open-source models are shown in Table~\ref{table:models}. The two proprietary models we evaluated, \texttt{gpt-3.5} and \texttt{gpt-4o}, are commercially available close-source models.

\begin{table}[h]
\centering
\resizebox{\columnwidth}{!}{%
\begin{tabular}{@{}ll@{}}
\toprule
\textbf{Models}        & \textbf{Sources}                                    \\ \midrule
\texttt{gpt-3.5-turbo} & \url{https://platform.openai.com/}                  \\
\texttt{gpt-4o}        & \url{https://platform.openai.com/}                  \\
\texttt{llama-3-70b}    & \url{https://huggingface.co/meta-llama/Meta-Llama-3-70B-Instruct}   \\
\texttt{qwen1.5-72b}   & \url{https://huggingface.co/Qwen/Qwen1.5-72B-Chat}  \\
\texttt{qwen1.5-110b}  & \url{https://huggingface.co/Qwen/Qwen1.5-110B-Chat} \\
\bottomrule
\end{tabular}%
}
\caption{
Sources of the evaluated models.
}
\label{table:models}
\end{table}

\end{document}